\documentclass[10pt,twocolumn,letterpaper]{article}

\usepackage[pagenumbers]{cvpr} 

\usepackage{graphicx}
\usepackage{amsmath}
\usepackage{amssymb}
\usepackage{booktabs}
\usepackage[linesnumbered, ruled,nofillcomment]{algorithm2e}
\usepackage{tabularx}
\usepackage{array}
\usepackage{multirow}
\usepackage{multicol}
\usepackage{xcolor}
\usepackage{xspace}
\usepackage{colortbl}
\usepackage{makecell}
\usepackage{float}
\usepackage{marvosym}

\makeatletter
\newcommand{\algorithmfootnote}[2][\footnotesize]{%
  \let\old@algocf@finish\@algocf@finish
  \def\@algocf@finish{\old@algocf@finish
    \leavevmode\rlap{\begin{minipage}{\linewidth}
    #1#2
    \end{minipage}}%
  }%
}
\makeatother

\usepackage[pagebackref,breaklinks,colorlinks]{hyperref}

\usepackage[capitalize]{cleveref}
\crefname{section}{Sec.}{Secs.}
\Crefname{section}{Section}{Sections}
\Crefname{table}{Table}{Tables}
\crefname{table}{Tab.}{Tabs.}
\Crefname{algorithm}{Algorithm}{Algorithms}
\crefname{algorithm}{Alg.}{Algs.}

\newcommand\blfootnote[1]{%
  \begingroup
  \renewcommand\thefootnote{}\footnote{#1}%
  \addtocounter{footnote}{-1}%
  \endgroup
}

\begin{document}

\title{SportsMOT: A Large Multi-Object Tracking Dataset in Multiple Sports Scenes}

\author
{
Yutao Cui$^{*}$ \quad
Chenkai Zeng$^{*}$ \quad
Xiaoyu Zhao$^{*}$ \quad
Yichun Yang$^{*}$ \quad
Gangshan Wu \quad
Limin Wang\textsuperscript{~\Letter}
\\[0.2cm]
State Key Laboratory for Novel Software Technology,
Nanjing University,
China
}
\maketitle

\blfootnote{* indicates equal contribution. \Letter~: Corresponding author.}

\begin{abstract}
Multi-object tracking in sports scenes plays a critical role in gathering players statistics, supporting further analysis, such as automatic tactical analysis. 
Yet existing MOT benchmarks cast little attention on the domain, limiting its development.
In this work, we present a new large-scale multi-object tracking dataset in diverse sports scenes, coined as \emph{SportsMOT}, where all players on the court are supposed to be tracked.
It consists of 240 video sequences, over 150K frames (almost 15× MOT17) and over 1.6M bounding boxes (3× MOT17) collected from 3 sports categories, including basketball, volleyball and football. 
Our dataset is characterized with two key properties: 1) fast and variable-speed motion and 2) similar yet distinguishable appearance. 
We expect SportsMOT to encourage the MOT trackers to promote in both motion-based association and appearance-based association.
We benchmark several state-of-the-art trackers and reveal the key challenge of SportsMOT lies in object association.
To alleviate the issue, we further propose a new multi-object tracking framework, termed as \emph{MixSort}, introducing a MixFormer-like structure as an auxiliary association model to prevailing tracking-by-detection trackers.
By integrating the customized appearance-based association with the original motion-based association, MixSort achieves state-of-the-art performance on SportsMOT and MOT17.
Based on MixSort, we give an in-depth analysis and provide some profound insights into SportsMOT.
The dataset and code will be available at \url{https://deeperaction.github.io/datasets/sportsmot.html}.  
\end{abstract}

\section{Introduction}
\label{sec:intro}

Multi-object tracking (MOT) has been a fundamental computer vision task for recent decades, aiming to locate the objects and associate them in video sequences. 
Researchers have cast much focus on various practical use cases like crowded street scenes~\cite{milan2016mot16,dendorfer2020mot20}, static dancing scenes~\cite{sun2022dancetrack} and driving scenarios~\cite{geiger2012we}, achieving considerable progress~\cite{bergmann2019tracking,bewley2016simple,wang2020towards,wang2021multiple,xu2021transcenter,peng2020chained,wu2021track} in MOT. 
MOT for sports scenes however is overlooked, where typically only the players on the court should be tracked for further analysis, such as counting the players' running distance or average speed and automatic tactical analysis.

\begin{figure*}[pt]
\centering
\includegraphics[width=0.95\linewidth]{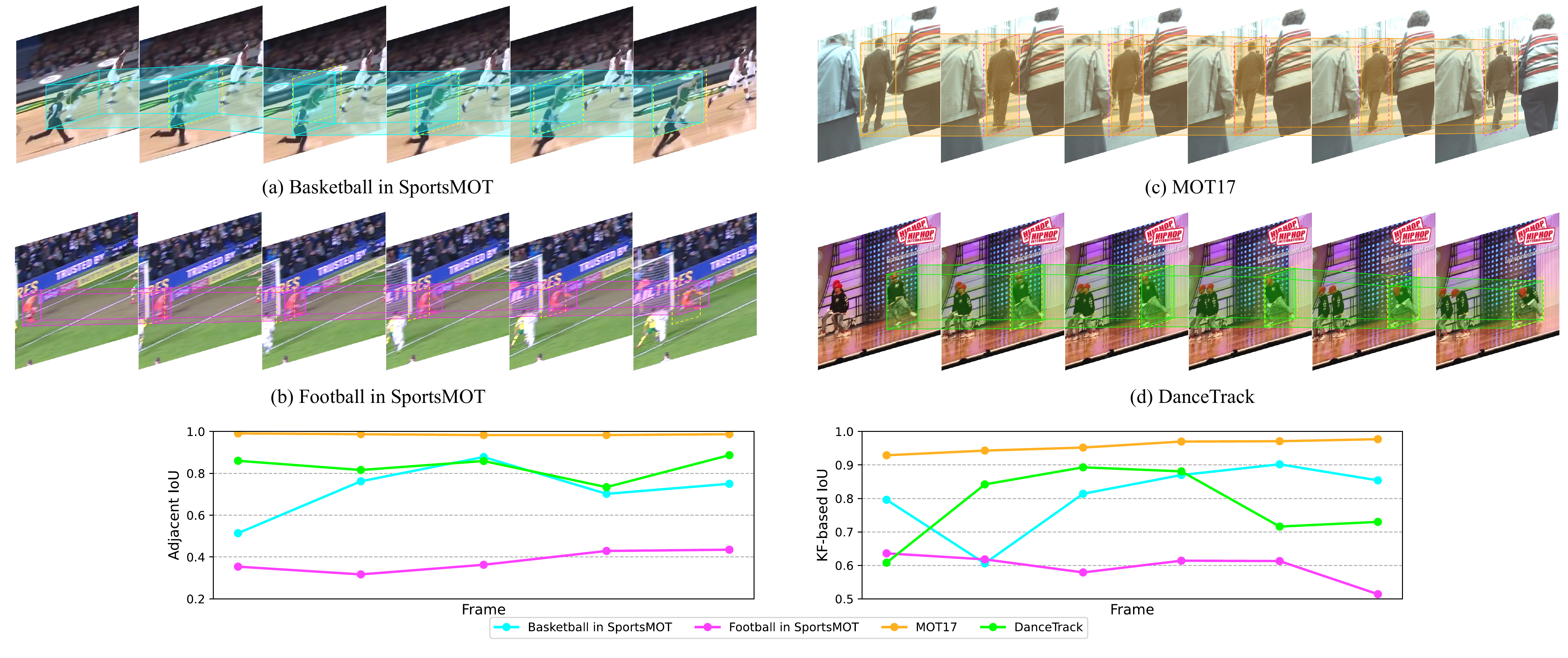}
\vspace{-3mm}
\caption{Sampled sequences from the categories of basketball and football of SportsMOT, MOT17 and DanceTrack. There exist two key properties of SportsMOT: 1) \textit{fast and variable-speed motion},~\ie players usually possess high speed and frequently change their running speed (the visualized adjacent IoU and Kalman Filter based adjacent IoU can indicate the property); 2) \textit{similar yet distinguishable appearance}, that is, players in sports scenes inherently wear jerseys with different numbers and usually display distinct postures. We expect SportsMOT to encourage the MOT trackers to promote in both motion-based association and appearance based association.}
\vspace{-5mm}
\label{fig:1}
\end{figure*}

Generally, prevailing state-of-the-art trackers~\cite{wojke2017simple,zhang2022bytetrack,cao2022observation,aharon2022bot,du2022strongsort} consist of several components to accomplish the tracking task: objects localization module, motion based objects association module and appearance based association module.
Biased to the data distribution of specific human tracking benchmarks, e.g. MOT17~\cite{milan2016mot16}, MOT20~\cite{dendorfer2020mot20} and DanceTrack~\cite{sun2022dancetrack}, the components of these trackers have difficulty adapting to sports scenes.
Firstly, motivated by surveillance or self-driving applications, current human tracking benchmarks provide tracks for almost all persons in the scenes. 
While for sports scenes like basketball or football games, generally only the players on the court are what we focus on, hence a specialized training platform is required to make the detectors suitable for sports scenes.
More importantly, in MOT17 and MOT20, these trackers highlight Kalman Filter~\cite{kalman1960new} based IoU matching for object association, due to the slow and regular motion of pedestrians.
DanceTrack highlights diverse motion rather than fast movement~\cite{sun2022dancetrack}, that is, dancers frequently switch the motion direction and relative position.
However in sports scenes, we observe \emph{fast and variable-speed movement} of objects on adjacent frames, \ie players usually possess high speed and frequently change their running speed in professional sports events, thus constituting barriers in existing motion based association.
For instance, as visualized in Fig.~\ref{fig:1}, the adjacent IoU and Kalman-Filter-based IoU in sports scenes remain lower than that on MOT17 and DanceTrack (More detailed comparison can be found in Fig.~\ref{fig:iou} and Fig.~\ref{fig:motion-iou}).
As a consequence, more suitable motion based association for sports scenes is required.
Additionally, compared to the MOT17 and MOT20 datasets in street scenes, the objects appearances in sports scenes are less distinguishable, since not only the players inherently are in similar clothes but also the players are frequently blurred caused by fast camera motion or targets motion.
Different from DanceTrack, where generally the dancers are in almost the same clothes and thus having indistinguishable appearance, players in sports scenes inherently wear jerseys with different numbers and usually display distinct postures.
Thereby, we argue that objects in sports scenes are with~\emph{similar yet distinguishable appearance}, which necessitates the appearance model developing more discriminative and extensive representations.

Considering the analysis above, to advance the development of tracking and sports analysis, we propose a multi-object tracking dataset in sports scenes, termed as \textbf{SportsMOT}. 
The dataset is large-scale, high-quality and contains dense annotations for every player on the court in various sports scenes.
It consists of 240 videos, over 150K frames (almost 15$\times$ MOT17~\cite{milan2016mot16}) and over 1.6M bounding boxes (3$\times$ MOT17~\cite{milan2016mot16}) collected from 3 categories of sports, including basketball, volleyball and football.
To provide a platform for making the trackers suitable for sports scenes, we split the dataset into training, validation and test subsets, consisting of 45, 45 and 150 video sequences respectively.
There exist two core properties of SportsMOT: (1) \textbf{fast and variable-speed motion}, requiring more suitable motion modeling association; (2) \textbf{similar yet distinguishable appearance}, which necessitates the appearance model developing more discriminative and extensive representations.
Altogether, we expect SportsMOT to encourage the trackers to promote in both the certain aspects, \ie motion based association and appearance based association.

Given the large-scale multi-object tracking dataset SportsMOT, we benchmark some recent tracking approaches and retrain all of them on the training split.
We observe IDF1 and AssA metrics are lower than that on MOT17 while the DetA is quite high, indicating that the main challenge of SportsMOT lies in objects association rather than objects localization.
To alleviate the issue, we propose a new multi-object tracking framework, dubbed as~\textbf{MixSort}, with introducing a MixFormer-like~\cite{cui2022mixformer} structure as appearance based association to prevailing tracking-by-detection trackers (\eg ByteTrack\cite{zhang2022bytetrack}, OC-SORT\cite{cao2022observation}).
By integrating the original motion based objects association and the designed appearance based association, the performance gets boosted on both SportsMOT and MOT17 benchmarks. 
Based on MixSort, we perform extensive exploration studies and provide some profound insights into SportsMOT.

The main contributions are summarized as follows:
\begin{itemize}
\vspace{-2.5mm}
\item We build a new large-scale multi-object tracking dataset in diverse sports scenes, SportsMOT, equipped with two key properties of 1) fast and variable-speed motion and 2) similar yet distinguishable appearance, aiming to advance the development of both tracking and sports analysis.
\vspace{-2.7mm}
\item We benchmark some prevailing trackers on SportsMOT, which reveals that the key challenge lies in objects association and hopefully can facilitate further research.
\vspace{-2.7mm}
\item We propose a new multi-object tracking framework MixSort, with introducing a MixFormer-like structure as appearance based association model to prevailing tracking-by-detection trackers, so as to boost the objects association.
Based on MixSort, we perform extensive studies and provide some profound insights into SportsMOT.
\end{itemize}

\section{Related Work}
\paragraph{Multi-object tracking datasets.}
Existing Multi-object tracking datasets usually focus on different scenes, such as autonomous driving, pedestrians on roads, and dancing. For autonomous driving, there are KITTI\cite{geiger2012we}, KITTI360\cite{liao2022kitti} and BDD100K\cite{yu2020bdd100k}, which focus on pedestrians and vehicles. Besides, other datasets, focusing on only the pedestrians, collect the videos from static and moving cameras. One of the earliest is PETS\cite{ferryman2009pets2009}, but it's too simple in some scenes. MOT15\cite{leal2015motchallenge} proposes the first large-scale benchmark for Multi-object tracking, followed by MOT17\cite{milan2016mot16} and MOT20\cite{dendorfer2020mot20}. It's worth noting that MOT20 focuses on extremely crowded scenes where many pedestrians are occluded, increasing the tracking difficulty greatly in both detection and association.
\begin{table}[pt]
\centering
\resizebox{\linewidth}{!}{
\begin{tabular}{l|ccccc}
\toprule
Dataset & Videos & Frames & Length (s) & Bbox & Tracks      \\
\midrule
MOT17   & 14 & 11,235 & 463 & 292,733 & 1,342 \\
MOT20   & 8  & 13,410 & 535 &  \textbf{1,652,040} & \textbf{3,456} \\
DanceTrack & 100 & 105,855 & 5,292 & - & 990 \\
SportsMOT  & \textbf{240} & \textbf{150,379} & \textbf{6,015} & 1,629,490 & 3,401  \\   
\bottomrule
\end{tabular}}
\vspace{-3mm}
\caption{Comparison of statistics between existing human MOT datasets and our SportsMOT.}
\vspace{-6mm}
\label{table_statistic}
\end{table}
Recently, DanceTrack\cite{sun2022dancetrack}, focusing on dancing scenes, is proposed to encourage trackers to rely less on visual discrimination and depend more on motion analysis. The emphasized properties are uniform appearance and diverse motion. While in SportsMOT, the appearance is similar yet distinguishable, and the players' motion is fast and with variable speed. We expect SportsMOT to encourage algorithms to promote in both appearance and motion association. Besides, SoccerNet~\cite{soccernet} is presented to track elements in football scenarios. The main difference lies in that, it only contains soccer scenes and tracks almost all elements (players, goalkeepers, referees, balls) on the court without distinction. While SportsMOT contains three types of sports where the objects in basketball scenes are more crowded and thus more challenging (refer to Section~\ref{exploration_sec}), and only focuses on the players to support further statistics and tactical analysis. 

\vspace{-4mm}
\paragraph{Object association in tracking.}
Association is a very important task in tracking, where trackers need to associate detections in new frames with existing tracks. For most of the trackers, a similarity matrix (or cost matrix) between new detections and tracks is computed based commonly on motion and appearance cues, which is later fed into Hungarian algorithm\cite{kuhn1955hungarian} to perform association. For example, SORT\cite{bewley2016simple} uses Kalman Filter to predict the location of objects and computes the IoU of detected and predicted bounding boxes as similarity matrix. IOU-Tracker\cite{bochinski2017high} directly computes the IoU without prediction. ByteTrack\cite{zhang2022bytetrack} adds an association phase for detection with low confidence score, which can boost the performance. OC-SORT\cite{cao2022observation} tries to address the limitation of Kalman Filter.

Appearance cues also play an important role in association. DeepSORT\cite{wojke2017simple} crops the detection from frame images, which are then used by networks to generate re-ID features. Then the motion cues and distance of re-ID features are fused to perform association. FairMOT\cite{zhang2021fairmot} uses a re-ID branch on a backbone shared by the detection branch to generate re-ID features. In CenterTrack\cite{zhou2020tracking}, the previous frame is used to help the prediction of tracks. Recently, Transformer\cite{vaswani2017attention} is used by some work such as TrackFormer\cite{meinhardt2022trackformer} and MOTR\cite{zeng2022motr} to boost the association quality.
Our proposed MixSort integrates the motion and appearance cues with a motion modeling component and the designed MixFormer-like structure respectively.

\section{SportsMOT Dataset}

\subsection{Dataset Construction}

\paragraph{Video Collection.} 
We select three worldwide famous sports, football, basketball, and volleyball, and collect videos of high-quality professional games including NCAA, Premier League, and Olympics from MultiSports~\cite{multisports}, which is a large dataset in sports area focusing on spatio-temporal action localization. 
Each category has typical players' formations and motion patterns, and they can effectively represent the diversity of sports scenarios.
Only the overhead shots of sports game scenes are used, guaranteeing certain extreme situations do not occur.
The proposed dataset consists of 240 video sequences in total, each of which is 720P and 25 FPS.
Following the principles of multi-object tracking, each video clip is manually checked to ensure that there are no abrupt viewpoint switches within the video.

\vspace{-2mm}
\paragraph{Annotation Pipeline.}
We annotate the collected videos according to the following guidelines.
\vspace{-2mm}
\begin{itemize}
  \item The entire athlete's limbs and torso, excluding any other objects like balls touching the athlete's body, are required to be annotated.
  \vspace{-2mm}
  \item The annotators are asked to predict the bounding box of the athlete in the case of occlusion, as long as the athletes have a visible part of body. However, if half of the athletes' torso is outside the view, annotators should just skip them.
  \vspace{-2mm}
  \item We ask the annotators to confirm that each player has a unique ID throughout the whole clip.
\end{itemize}

We provide a customized labeling tool for SportsMOT and a corresponding manual book to annotators.
Once they start annotating a new object, the labeling tool automatically assigns a new ID to the object and propagates the bounding box of previous state to the current state, with the help of the single object tracker KCF~\cite{hannuna2019ds}.
Then the generated bounding boxes should be refined by the annotators, so as to improve annotation quality.
After carefully reviewing each annotation result, we refine the bounding boxes and IDs that do not satisfy the standards, hence building a high-quality dataset.
Finally, the bounding boxes with too small size, ~\ie $ w<5 $ or $ h < 5 $, are deleted.

\subsection{Dataset Statistic}
\label{sec:dataset}
\paragraph{Overview.} 
SportsMOT is a large-scale and high-quality MOT dataset, aiming to advance the development of both sports analysis and multi-object tracking.
Table~\ref{table_statistic} compares the statistics of SportsMOT with the prevailing human tracking datasets, including MOT17, MOT20 and DanceTrack.
According to the statistics, SportsMOT has a large number of bounding boxes of over 1.6M, which is comparable to MOT20 and significantly larger than MOT17.
Besides, SportsMOT has a large number of video clips, tracks (2.5$\times$ MOT17, 3.4$\times$ DanceTrack), frames (13.4 $\times$ MOT17).
As shown in Table~\ref{table_categary}, we also compare the basic statistics of each category of SportsMOT.
SportsMOT solely provides fine annotations of the players on the court, which are supposed to be tracked for further analysis.
To provide a platform for making the trackers suitable for sports scenes, we split the dataset into training, validation and test subsets, consisting of 45, 45 and 150 video sequences respectively.

\begin{figure}[pt]
\centering
\includegraphics[width=1.0\linewidth]{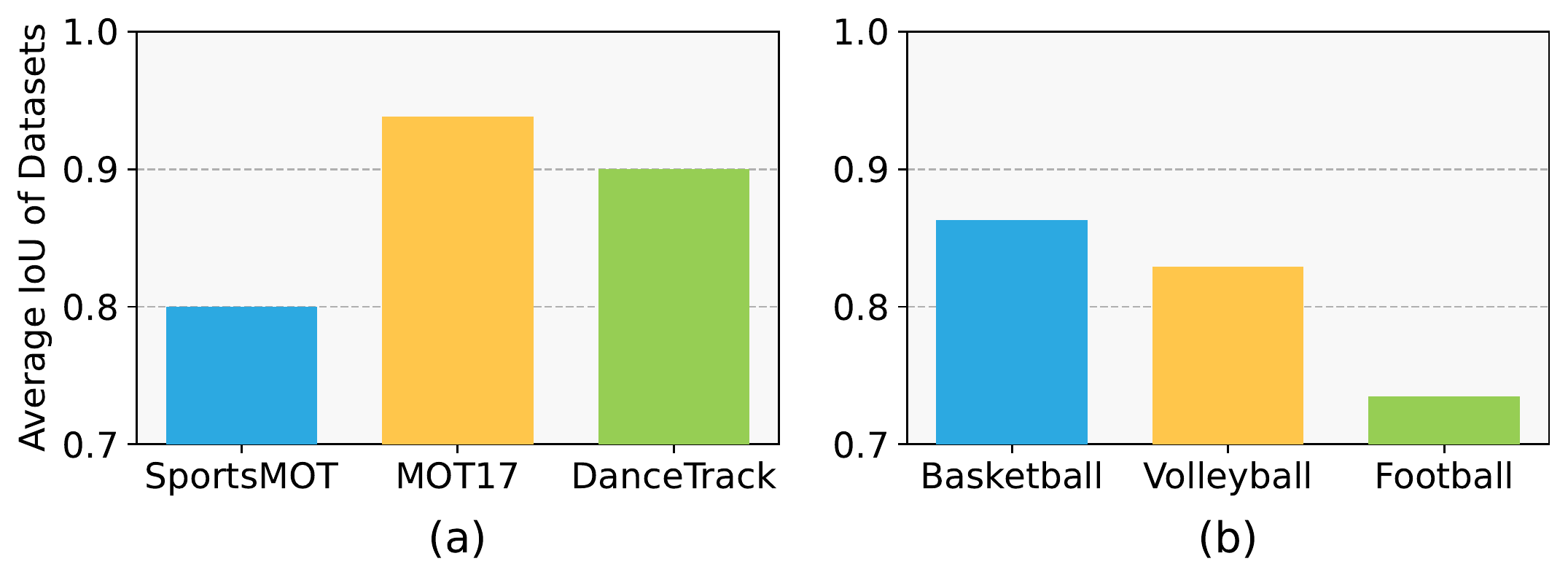}
\vspace{-8mm}
\caption{IoU on adjacent frames. (a) Compared to MOT17 and DanceTrack, SportsMOT has a lower score, indicating that objects have faster motion. (b) In SportsMOT, the category of football has the lowest IoU score, which means that football players often have fast motion.}
\label{fig:iou}
\end{figure}

\begin{table}[pt]
 \centering
\resizebox{\linewidth}{!}{
\begin{tabular}{l|ccccc}
\toprule
\multirow{2}{*}{Category} & \multirow{2}{*}{Frames} & \multirow{2}{*}{Tracks} & Track & Track  & Bboxes \cr
&  &  & gap len. & len. & per frame
\\
\midrule
Basketball  & 845.4 & 10  & 68.7  & 767.9 & 9.1  \\
Volleyball  & 360.4 & 12  & 38.2  & 335.9 & 11.2 \\
Football    & 673.9 & 20.5  & 116.1  & 422.1 & 12.8 \\
Total       & 626.6 & 14.2  & 96.6   & 479.1 & 10.8  \\      \bottomrule           
\end{tabular}}
\vspace{-2mm}
\caption{Detailed statistics of the three categories in SportsMOT.}
\vspace{-6mm}
\label{table_categary}
\end{table}

\begin{figure}[pt]
\centering
\includegraphics[width=1.0\linewidth]{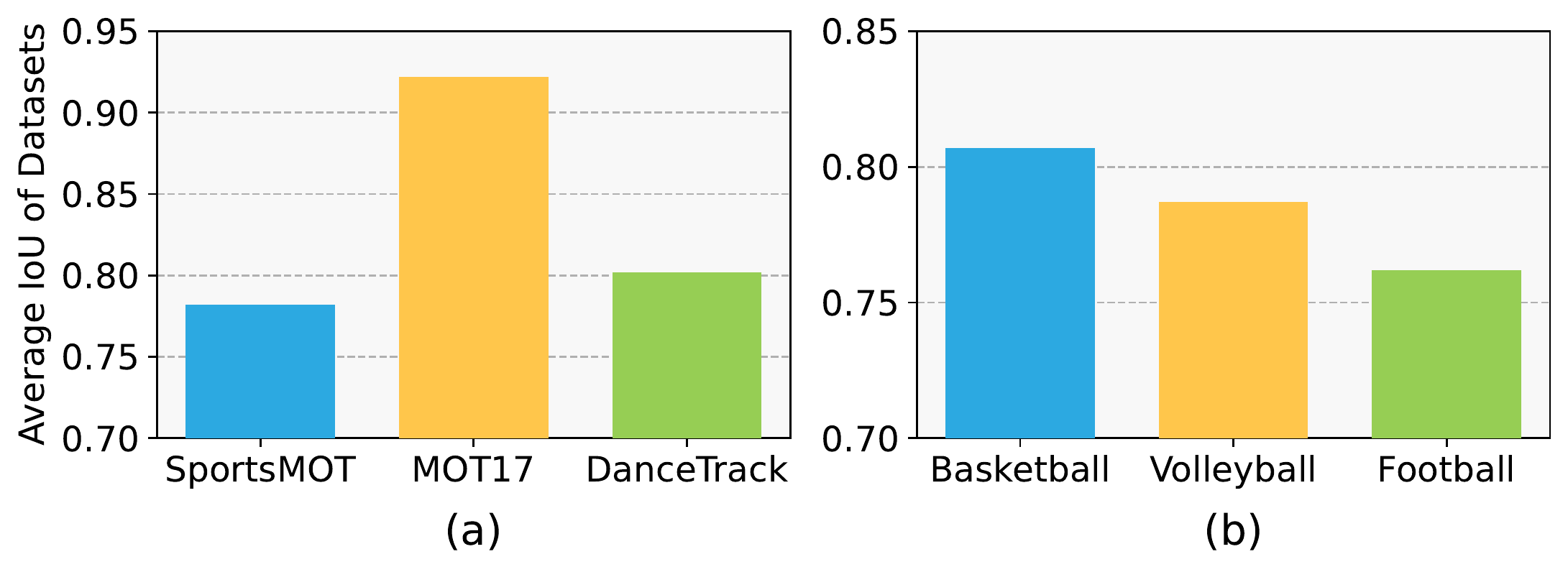}
\vspace{-8mm}
\caption{Kalman-Filter-based IoU on adjacent frames. (a) Compared to MOT17 and DanceTrack, SportsMOT has a lower score, indicating that objects have more variable-speed motion. (b) In SportsMOT, the category of football has the lowest Kalman-Filter-based IoU score.}
\vspace{-2mm}
\label{fig:motion-iou}
\end{figure}

\vspace{-4mm}
\paragraph{Fast and Variable-Speed Motion.}
Motion cues play an important role in object association for multi-object tracking.
The existing human-tracking datasets (\eg MOT17, MOT20 and DanceTrack) generally have certain motion patterns that are distinct with sports scenes, constituting barriers in players tracking.
For instance, in MOT17 and MOT20, pedestrians are featured by linear motion with constant speed, which is easily hacked by association strategies with constant velocity assumption.
Besides, DanceTrack highlights diverse motion rather than fast motion, that is, dancers usually move in more diverse directions with relatively low speed.
In contrast, SportsMOT has distinct motion patterns,~\ie \emph{fast and variable-speed motion}, where the players typically move fast with their running speed or camera speed frequently changing.
As illustrated in Fig.~\ref{fig:iou}, among the three datasets, SportsMOT has the lowest IoU score of the objects bounding boxes on adjacent frames, indicating the fast movement.
We use the ground truth of previous frames for Kalman Filter prediction. The result and current ground truth are used to calculate Kalman-Filter-based IoU.
Seen from Fig.~\ref{fig:motion-iou}, SportsMOT also has the lowest Kalman-Filter-based IoU score on adjacent frames, which suggests that the motion can not be easily modeled by prevailing methods due to the variable-speed movement.
Specifically, football has the smallest adjacent IoU and Kalman-Filter-based IoU, which is closely related to the fast running speed, abrupt acceleration or stops.
It poses a major challenge for trackers based on simple motion assumptions and also encourages them to model object motion in more dynamic and adaptive ways.

\begin{figure}[pt]
	\centering
	\begin{subfigure}{0.32\linewidth}
		\centering
		\includegraphics[width=\linewidth]{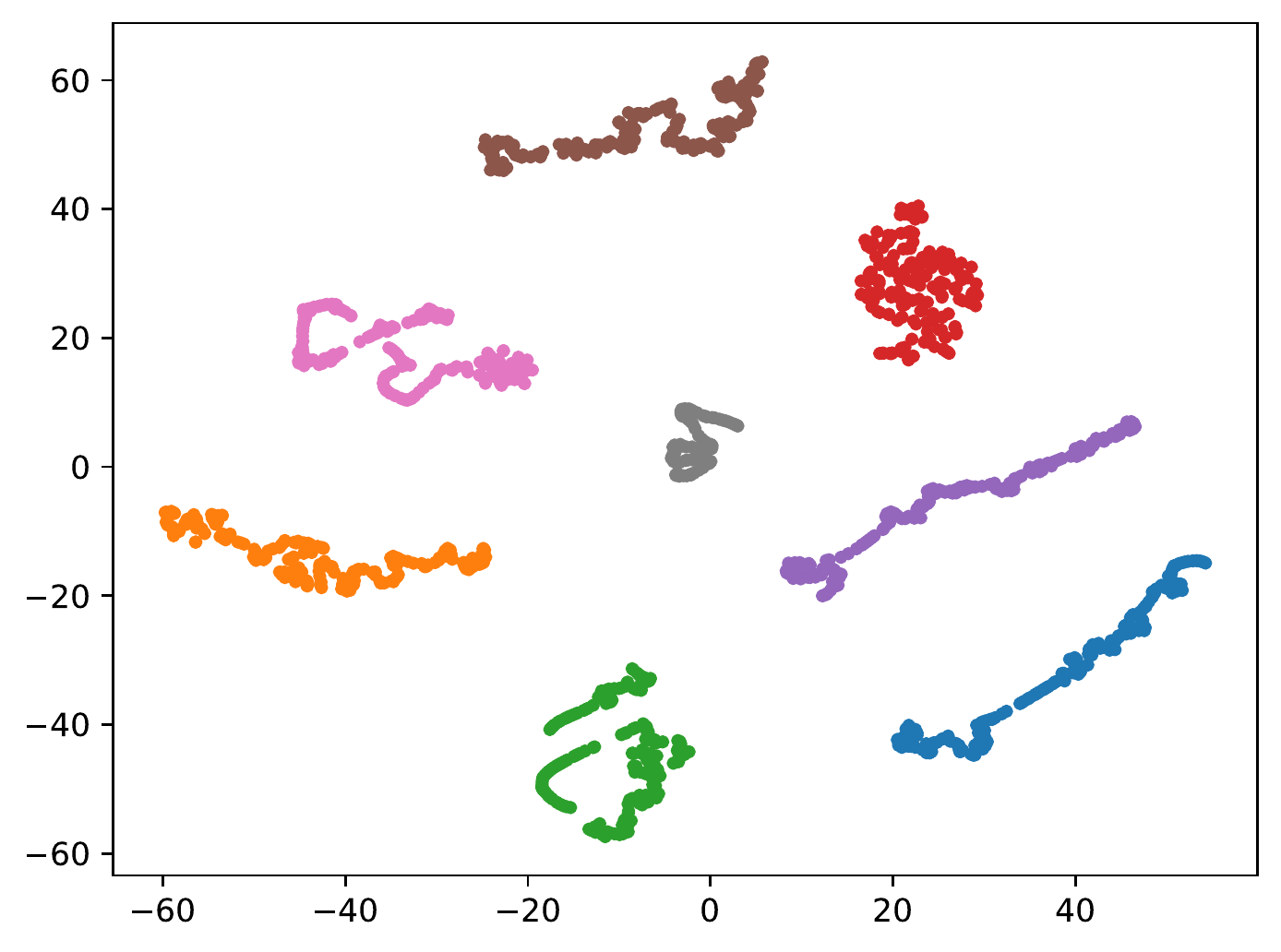}
		\caption{MOT17}
		\label{reid:mot17}
	\end{subfigure}
 \begin{subfigure}{0.32\linewidth}
        \centering
        \includegraphics[width=\linewidth]{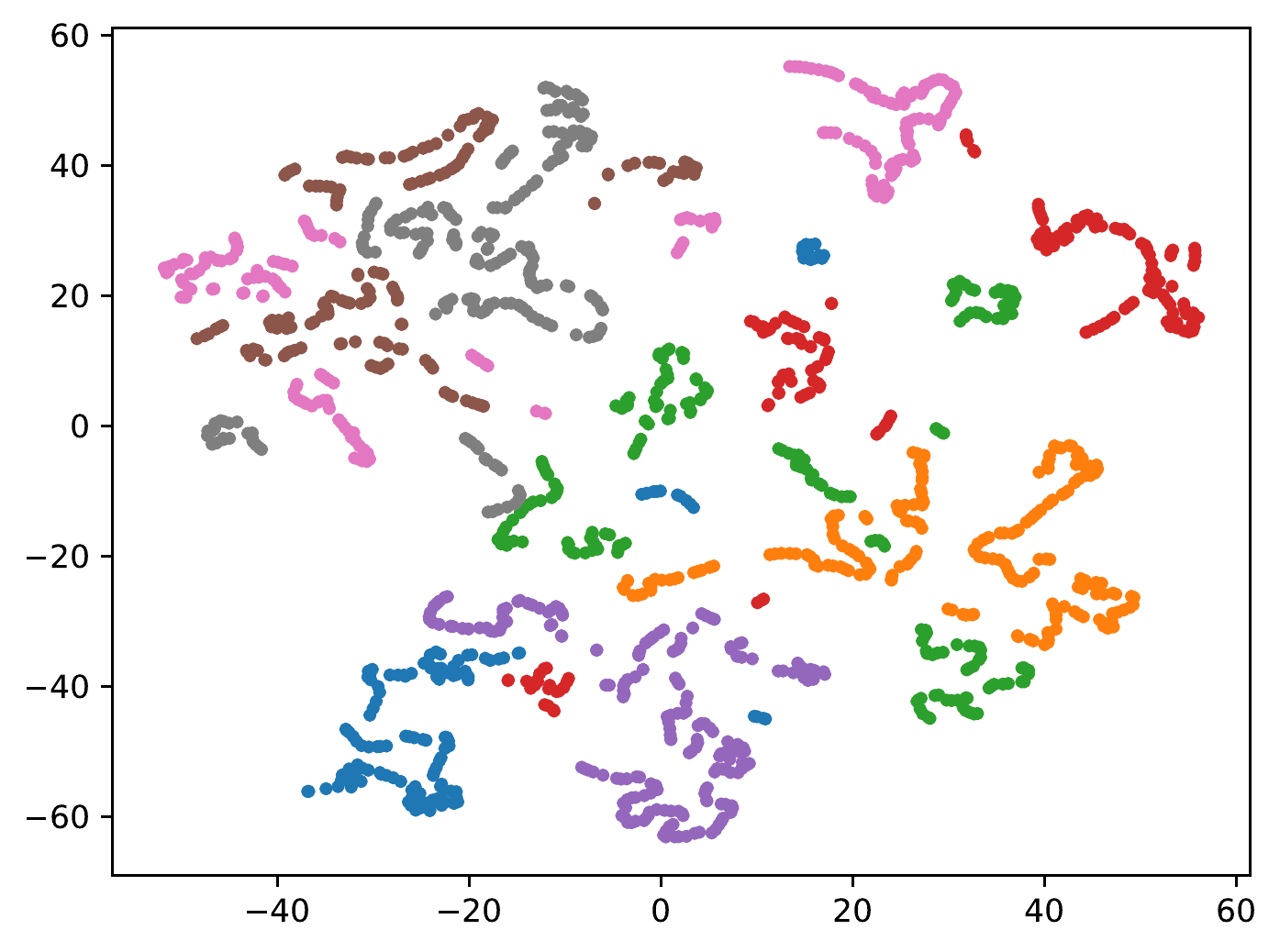}
        \caption{SportsMOT}
        \label{reid:sports}
    \end{subfigure}
	\begin{subfigure}{0.32\linewidth}
		\centering
		\includegraphics[width=\linewidth]{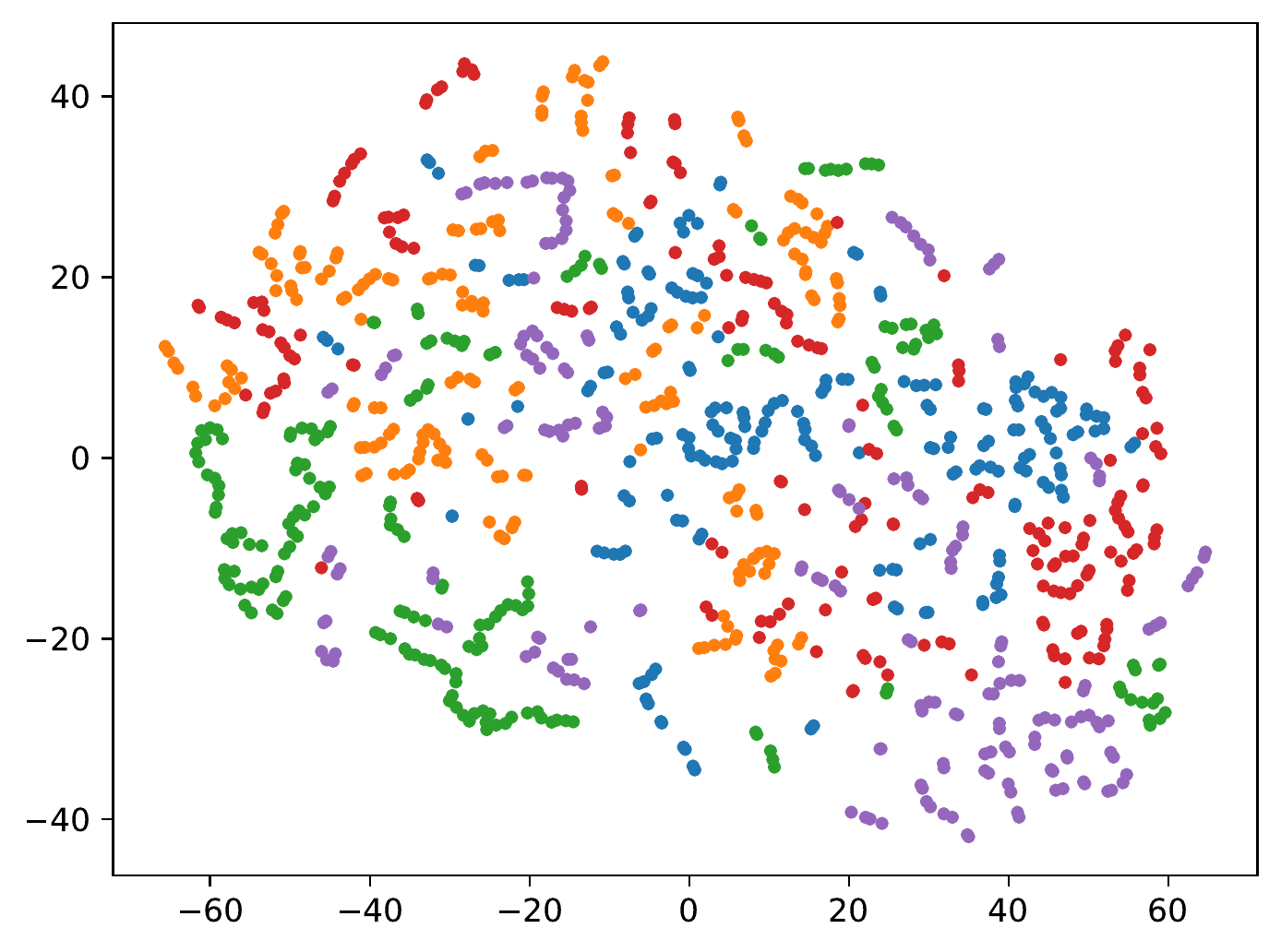}
		\caption{DanceTrack}
		\label{reid:dancetrack}
	\end{subfigure}
\vspace{-2mm}
\caption{Visualization of re-ID features from sampled videos
in MOT17, SportsMOT and DanceTrack dataset using t-SNE~\cite{van2008visualizing}.
The same object is coded by the same color. It indicates that object appearance of SportsMOT is less distinguishable than that of MOT17, while more distinguishable than that of DanceTrack. We expect the appearance model to capture more discriminative and extensive representation for object association.
}
\vspace{-5mm}
\label{fig:reid}
\end{figure}

\vspace{-4mm}
\paragraph{Similar yet Distinguishable Appearance.}
Object appearance is another kind of cue on which MOT trackers often rely to distinguish different objects.
In MOT17 and MOT20, pedestrians are usually distinct in body size and wear different clothes, yielding discriminative visual features.
In contrast, the objects in DanceTrack typically wear nearly identical outfits, leading to indistinguishable appearances.
Thereby, DanceTrack highlights solely relying on motion-based association rather than appearance-based association.
In SportsMOT, the players also have very similar appearances.
However, the players wear jerseys with different numbers and usually display distinct postures, thus resulting in~\emph{similar yet distinguishable} appearance.
In Fig.~\ref{fig:reid}, we provide visualization of re-ID feature from sampled videos in MOT17, SportsMOT and DanceTrack dataset using t-SNE~\cite{van2008visualizing}. 
It implies that the re-ID features in SportsMOT are similar yet distinguishable compared to MOT17 and DanceTrack. 
We aim to encourage the trackers to learn more discriminative visual representations for more robust object association.

\subsection{Evaluation Metrics}

MOTA~\cite{bernardin2008evaluating} is the main metric for existing MOT evaluations. However, MOTA focuses more on measuring the accuracy of detection. 
To highlight the performance of object association, we recommend HOTA~\cite{luiten2021hota}, AssA and IDF1~\cite{ristani2016performance} as the major evaluation metrics in SportsMOT dataset.
HOTA aims to measure the accuracy of detection and association equally and has also been found to be more consistent with human intuition.

\begin{figure}[pt]
\centering
\includegraphics[width=1.0\linewidth]{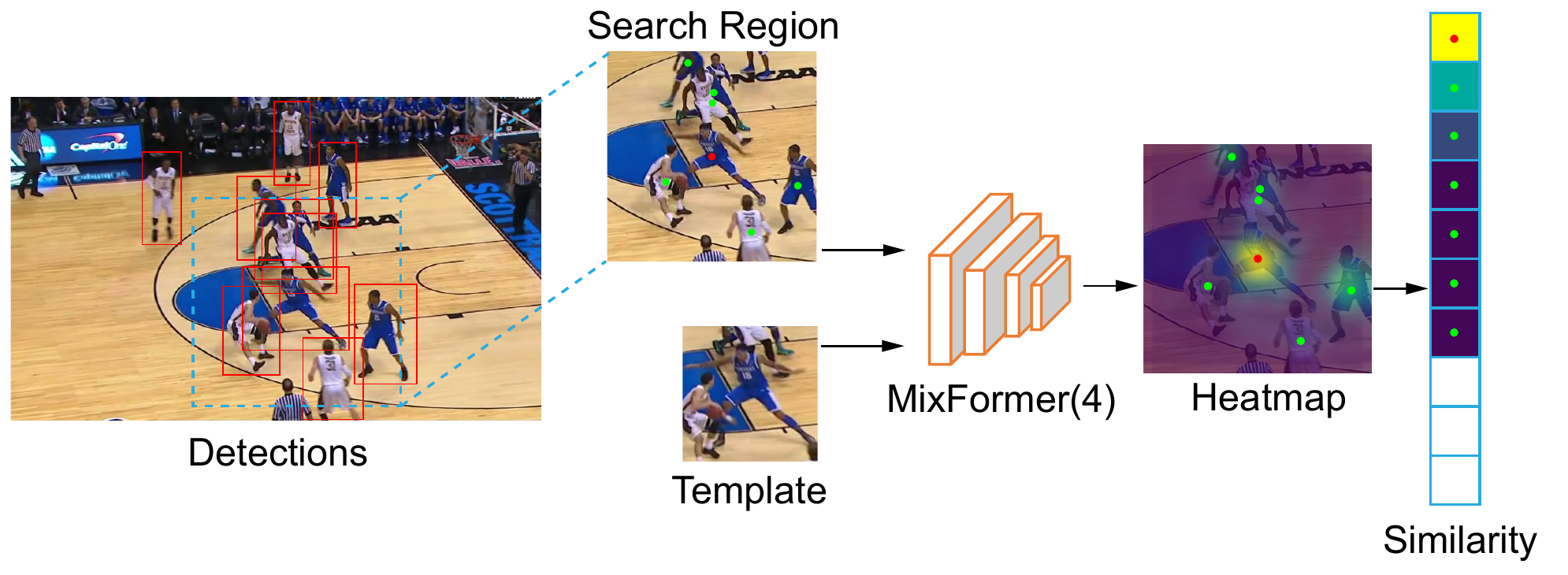}
\vspace{-7mm}
\caption{Paradigm for computing visual similarity matrix of tracks and detections in MixSort. The center of ground-truth target detection is marked with red dots, the others are green. The blue dashed box indicates the cropped search region. The blank part of similarity vector means that for detections not in the search region, the corresponding value is set to 0.}
\vspace{-4mm}
\label{fig:2}
\end{figure}

\section{Multi-Object Tracking on SportsMOT}
In this section, we present our proposed multi-object tracking framework, called \textit{MixSort}. This framework is designed to enhance the appearance-based association performance and can be applied to any trackers that follow the tracking-by-detection paradigm, such as ByteTrack\cite{zhang2022bytetrack} and OC-SORT\cite{cao2022observation}.

We begin by explaining how we use the MixFormer\cite{cui2022mixformer} network to compute visual similarities between tracked templates and detected objects in multi-object tracking. Next, we describe the overall pipeline of MixSort. Finally, we provide details on the training and inference of MixSort.

\subsection{MixFormer for Appearance-based Association}
\paragraph{MixFormer.} 
In this paragraph, we discuss the use of MixFormer in our proposed framework \textit{MixSort}. MixFormer is designed to extract target-specific discriminative features and perform extensive communication between the target and search area, therefore, it is the key component that enables MixSort to compute visual similarities between the templates of tracked objects and detected objects in the search region of the current frames.

The original MixFormer uses a corner-based localization head to predict the top-left and bottom-right corners of the input template in the search region. However, we modify the corner head by using a heatmap prediction head that predicts the center of the template and generates a confidence heatmap. This allows us to compute the similarity between the detection and the template.

To make MixSort suitable for multi-object tracking and accelerate inference speed, we reduce the number of mixed attention modules in MixFormer from 12 to 4. The steps involved in computing the visual similarity matrix are illustrated in \Cref{fig:2}.

\vspace{-4mm}
\paragraph{Association Strategy.} 
In order to perform association between detections and existing tracks, we use a mixed similarity matrix generated by computing the visual similarity between the target template and the detected objects in the search region of the current frame. Specifically, we obtain the heatmap response at the center of each detection as its visual similarity to the template. The resulting similarity matrix is then combined with the IoU matrix using the Hungarian Algorithm.

To start, for each existing track $t$, we use the Kalman Filter to predict its new location. Then, we crop the current frame centered at the predicted location with a certain scale to obtain the search region $s$. By feeding $s$ and the template $t$ into MixFormer, we generate a heatmap $\mathcal{H}$ that represents the similarity between the template and search region.

Next, for each detection $d$ whose center is in the search region $s$, we set its similarity to track $t$ as the response in the heatmap $\mathcal{H}$; the similarity values of other detections are set to $0$. Finally, we fuse the visual similarity and the IoU score to obtain the mixed similarity matrix
\begin{equation}
	M=\alpha\cdot\text{IoU}+(1-\alpha)\cdot V
    \label{eq:M}
\end{equation}
where $\alpha$ is the weight coefficient and $V$ represents the visual similarity matrix calculated using MixFormer.

\subsection{MixSort Tracking}
\begin{figure}[pt]
\centering
\includegraphics[width=1.0\linewidth]{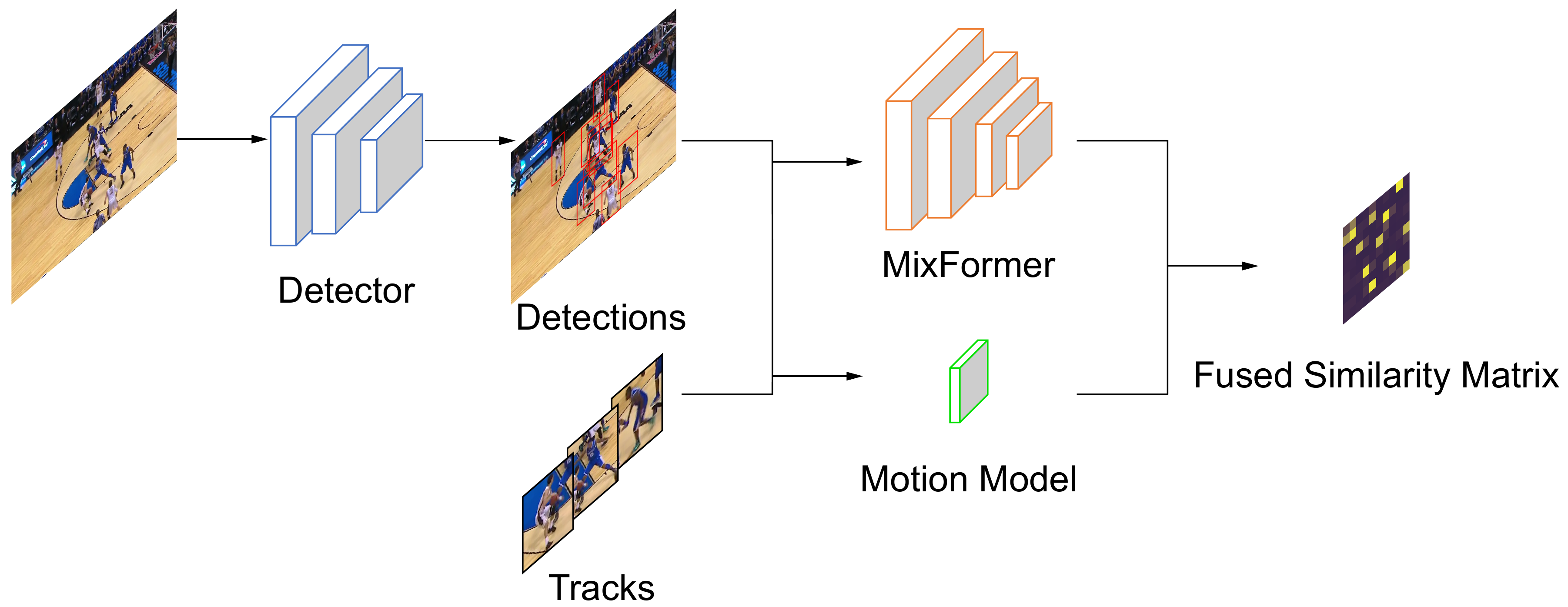}
\vspace{-5mm}
\caption{The pipeline of MixSort. We use motion model and MixFormer to generate fused similarity matrix for association.}
\vspace{-5mm}
\label{fig:3}
\end{figure}

Based on the tracking-by-detection paradigm, the pipeline of \textit{MixSort} can be generalized as follows:

As shown in \Cref{fig:3}, we first obtain detections using an object detector. Then, we employ a motion model (e.g., Kalman Filter) to predict new locations of existing tracks. Based on the new locations and templates of tracks, we compute a fused similarity matrix as described above and use it to associate tracks and detections by the Hungarian Algorithm. Finally, for matched tracks and detections, we update the online templates. For unmatched tracks, we keep them until the threshold is reached. For unmatched detections with confidence scores higher than the threshold, we initialize new tracks.

\subsection{Training and Inference}
\paragraph{Training.} We only consider the training of MixFormer here since the detector remains the same as initial method (\eg ByteTrack). The original MixFormer is trained on SOT datasets, so we first modify the format of ground truth of MOT datasets, that is, converting the ground truth trajectory of every single player into TrackingNet format~\cite{trackingnet}.\\
\indent For each ground-truth bounding box, we compute its corresponding center location $(c_x,c_y)$ in the low-resolution heatmap. Following CornerNet\cite{law2018cornernet}, the ground-truth heatmap response is generated using 2D Gaussian kernel:
\begin{equation}
	h_{xy}=\exp (-\frac{(x-c_x)^2+(y-c_y)^2}{2\sigma^2})
\end{equation}
where $\sigma$ is adaptive to the size of the bounding box. The training loss is a pixel-wise logistic regression with focal loss\cite{lin2017focal}:
\begin{equation}
	L=-\sum_{xy}\begin{cases}
		(1-\hat h_{xy})^\gamma\log(\hat h_{xy})&, h_{xy}=1;\\
		(1-h_{xy})^\beta(\hat h_{xy})^\gamma\log(1-\hat h_{xy})&,\text{otherwise.}
	\end{cases}
\end{equation}
where $\gamma$ and $\beta$ are hyper-parameters in focal loss, and we set $\gamma=2, \ \beta=4$ following CornerNet.

\vspace{-4mm}
\paragraph{Inference.}
For each track, we maintain only one template for keeping a balance between accuracy and speed. When a detection is matched to an existing track, we directly replace the original template to the new detection, if and only if the ratio of its uncovered (\ie overlapping with any detected objects) area is larger than a certain threshold, so as to reduce the impact of misleading representations.

\section{Experiments and Analysis}

\subsection{Experiment Setup}
\paragraph{Dataset Split.}
In benchmark experiments, we follow the default split described in \cref{sec:dataset}. In exploration study, we split the original MOT17 training set into two sets, used for training and validation respectively following CenterTrack.

\vspace{-4mm}
\paragraph{Implementation Details.}
Following ByteTrack and OC-SORT, we use YOLOX~\cite{ge2021yolox} as our detector. Using COCO-pretrained model as the initialized weights, we first train the model on CrowdHuman~\cite{shao2018crowdhuman} for 80 epochs and then train on SportsMOT for another 80 epochs. The remaining settings are the same as that in ByteTrack.

For MixFormer, we initialize the backbone with the model trained on VOT datasets and then fine-tune it on SportsMOT for 300 epochs with learning rate initialized as $1e-4$ and decreased to $1e-5$ at epoch 200. The optimizer is ADAM\cite{kingma2014adam} with weight decay $10^{-4}$. The sizes of search images and templates are $224\times224$ and $96\times96$ respectively. The max sample interval is set to 10.
For every tracking result, we apply linear interpolation as post-processing, with maximum gap set to 20.

\subsection{Benchmark Results}

\begin{table*}[!t]
\begin{center}{
\resizebox{\linewidth}{!}{
\begin{tabular}{l|c|cccccccc}
\toprule

&Training Setup&HOTA$\uparrow$ & IDF1$\uparrow$ & AssA$\uparrow$ & MOTA$\uparrow$ & DetA$\uparrow$& LocA$\uparrow$ & IDs$\downarrow$  & Frag$\downarrow$  \\ 

\midrule
CenterTrack~\cite{zhou2020tracking} & Train           & 62.7 & 60.0 & 48.0 & 90.8 & 82.1& 90.8 & 10481  & 5750 \\
FairMOT~\cite{zhang2021fairmot}       & Train           & 49.3 & 53.5 & 34.7 & 86.4  & 70.2& 83.9 & 9928  & 21673 \\
QDTrack~\cite{pang2021quasi}      & Train          & 60.4 & 62.3 & 47.2 & 90.1 & 77.5 & 88.0 & 6377  & 11850 \\
TransTrack~\cite{sun2020transtrack}    & Train         & 68.9 & 71.5 & 57.5 & 92.6  & 82.7 & 91.0 & 4992  & 9994 \\
GTR~\cite{zhou2022global}      & Train             & 54.5 & 55.8 & 45.9  & 67.9 & 64.8 & 89.0 & 9567  &  14525 \\ 
ByteTrack~\cite{zhang2022bytetrack}    & Train           & 62.8 & 69.8 & 51.2 & 94.1 & 77.1 & 85.6 & 3267 &  4499 \\
OC-SORT~\cite{cao2022observation}   & Train            & 71.9 & 72.2 & 59.8 & 94.5 & 86.4 & 92.4 & 3093 &  3474 \\
ByteTrack & Train+Val & 64.1 & 71.4& 52.3 & 95.9  & 78.5 & 85.7 &  3089 & 4216\\
OC-SORT & Train+Val & 73.7 & 74.0 & 61.5 & 96.5 & 88.5 & 92.7 & 2728 & \textbf{3144} \\
\midrule
MixSort-Byte  & Train+Val    & 65.7~\textcolor{green}{(+1.6)} & 74.1~\textcolor{green}{(+2.7)}& 54.8~\textcolor{green}{(+2.5)} & 96.2  & 78.8 & 85.7 & \textbf{2472} &  4009 \\
MixSort-OC    & Train+Val    & \textbf{74.1}~\textcolor{green}{(+0.4)} & \textbf{74.4}~\textcolor{green}{(+0.4)}  & \textbf{62.0}~\textcolor{green}{(+0.5)} & \textbf{96.5} & \textbf{88.5} & \textbf{92.7}& 2781  & 3199  \\
\bottomrule
\end{tabular}}}
\end{center}
\vspace{-6mm}
\caption{Tracking performance of investigated algorithms on our proposed SportsMOT. The best results are shown in~\textbf{bold}.}
\vspace{-4mm}
\label{table_mot}
\end{table*}

We evaluate several representative methods of three kinds on our dataset. ByteTrack\cite{zhang2022bytetrack}, OC-SORT\cite{cao2022observation} and QDTrack\cite{pang2021quasi} are trackers in tracking-by-detection paradigm. CenterTrack\cite{zhou2020tracking} and FairMOT\cite{zhang2021fairmot} perform joint detection and tracking in one stage. TransTrack\cite{sun2020transtrack} and GTR\cite{zhou2022global} are trackers based on Transformer. Most of the current best multi-object tracking algorithms belong to tracking-by-detection paradigm, however, due to the separation of detection and tracking, the information cannot be shared completely. Joint-detection-and-tracking paradigm couples the two modules, with the goal of boosting the performance of each. Transformer-based-tracking methods are relatively new but have achieved great performance. Despite its huge potential, the model complexity and calculation cost are much higher, resulting in large memory and long training time.

All training settings including the number of epochs and change of learning rate are consistent with original papers. According to different default settings, we follow the commonly used pretraining datasets, such as CrowdHuman\cite{shao2018crowdhuman}, COCO\cite{lin2014microsoft} and ImageNet\cite{deng2009imagenet}, and apply SportsMOT-train with or without other datasets for finetuning for different methods. We compare the results in \cref{table_mot}.

In sports scenes, the clear appearance and sparse density of objects allow current mature detection frameworks to generate bounding boxes with high accuracy. However, specialized detectors need to be trained for not detecting audience and referees. The key challenges are fast speed and motion blur, which forces us to pay more attention to improve the association performance. Besides, The wide range of HOTA and MOTA denotes SportsMOT is more distinguishable among different kinds of algorithms. 

Tracking-by-detection paradigm methods like ByteTrack and OC-SORT outperform most of the methods in the table. But their association performance is still not satisfactory enough. Thus we propose MixSort that can be applied to any trackers following this paradigm and achieve state-of-the-art performance on SportsMOT.
Besides, to further validate the effectiveness of MixSort, we compare MixSort with the state-of-the-art trackers on MOT17 validation set and test set under the private detection protocol in \cref{fig:mot17testval}. Our MixSort-byte and MixSort-OC outperform these trackers in HOTA, IDF1 and AssA metrics.

\subsection{Exploration Study}
\label{exploration_sec}
In this section, we perform extensive studies on the proposed MixSort and SportsMOT. 

\vspace{-4mm}
\paragraph{Effectiveness of the proposed association module.}
We evaluate the effectiveness of MixSort by applying it to two state-of-the-art trackers, OC-SORT\cite{cao2022observation} and ByteTrack\cite{zhang2022bytetrack}, which follow the tracking-by-detection paradigm and use YOLOX as their detector. The evaluation is conducted on the SportsMOT test set, and the results are presented in \cref{table_mot}. Our experiments show that MixSort significantly improves the performance of both trackers, with OC-SORT achieving a 0.4 HOTA increase and ByteTrack achieving a 1.6 HOTA increase on SportsMOT. This demonstrates the effectiveness of MixSort in enhancing the association.

\vspace{-4mm}
\paragraph{Appearance-based \textit{vs}. Motion-based association.}
\begin{table}[pt]
\centering
\resizebox{\linewidth}{!}{
\begin{tabular}{@{}l|ccccccccccc@{}}
\toprule
$\alpha$   & 1    & 0.9  & 0.8  & 0.7  & 0.6  & 0.5  & 0.4  & 0.3  & 0.2  & 0.1  & 0    \\ \midrule
basketball & 65.9 & 66.1 & \textbf{66.2} & 65.9 & 65.3 & 64.8 & 63.6 & 60.7 & 56.7 & 47.1 & 26.9 \\
volleyball & 76.0 & 76.8 & 76.5 & 76.4 & 76.5 & \textbf{76.9} & 76.4 & 75.5 & 73.7 & 69.1 & 40.9 \\
football   & 71.9 & 72.4 & 72.3 & 72.4 & 72.5 & 72.8 & \textbf{73.2} & 72.9 & 72.6 & 71.4 & 65.7 \\ \bottomrule
\end{tabular}}
\vspace{-2mm}
\caption {Comparison of the HOTA metric of basketball, volley and football under different fusion parameters $\alpha$ on SportsMOT test set. The models are trained on SportsMOT training set.}
\label{tab:alpha}
\vspace{-6mm}
\end{table}
We have demonstrated that MixSort can improve association performance. 
In this paragraph, we investigate the impact of the fusion weight $\alpha$ on the ability of appearance cues to aid in conventional motion-based association. We evaluate OC-SORT with MixSort on the three categories in the SportsMOT test set using $\alpha$ values ranging from 1 to 0 in \cref{eq:M}. The results, presented in \cref{tab:alpha}, reveal that pure motion-based association ($\alpha=1$) outperforms pure appearance-based association ($\alpha=0$) in all categories, underscoring the significance of motion cues in sports scenes. Moreover, fused association surpasses both pure motion-based and appearance-based association, suggesting that both motion and appearance cues should be considered jointly for optimal results.

Our analysis of the three categories reveals that appearance cues provide the most significant improvement for football videos (+1.3), followed by volleyball (+0.9) and basketball (+0.3). Combining this with \cref{fig:iou}, which indicates that the adjacent IoU of football games is the smallest (i.e., motion is fastest) among the three categories, we can conclude that scenes with faster motion are more dependent on appearance cues.

\begin{table*}[ht]
\resizebox{\linewidth}{!}{
\begin{tabular}{l|cccccc|cccccc}
\toprule
\multicolumn{1}{l|}{\multirow{2}{*}{}} & \multicolumn{6}{c|}{MOT17-test} & \multicolumn6{c}{MOT17-val} \\
\multicolumn{1}{l|}{} & HOTA$\uparrow$ & IDF1$\uparrow$ & AssA$\uparrow$ & MOTA$\uparrow$ & DetA$\uparrow$ & IDs$\downarrow$ & HOTA$\uparrow$ & IDF1$\uparrow$ & AssA$\uparrow$ & MOTA$\uparrow$ & DetA$\uparrow$ & IDs$\downarrow$\\
\midrule
QDTrack~\cite{pang2021quasi} & 53.9 & 66.3 & 52.7 & 68.7 & 55.6 & 3378 & - & - & - & - & - & -\\
MOTR~\cite{zeng2022motr} & 57.2 & 68.4 & 55.8 & 71.9 & 58.9 & 2115 & - & - & - & - & - & -\\
GTR~\cite{zhou2022global} & 59.1 & 71.5 & 57.0 & 75.3 & 61.6 & 2859 & 63.0 & 75.9 & 66.2 & 71.3 & 60.4 & -\\
ByteTrack~\cite{zhang2022bytetrack} & 63.1 & 77.3 & 62.0 & \textbf{80.3} & \textbf{64.5} & 2196 & - & 79.7 & - & 76.7 & - & 159\\
OC-SORT~\cite{cao2022observation} & 63.2 & 77.5 & 63.4 & 78.0 & 63.2 & 1950 & 68.0 & 79.3 & 69.9 & 77.9 & - & -\\
\midrule
MixSort-Byte & \textbf{64.0} & \textbf{78.7} & \textbf{64.2} & 79.3 & 64.1 & 2235 & \textbf{69.4} & \textbf{81.1} & 71.3 & \textbf{79.9} & \textbf{68.2} & 155\\
MixSort-OC & 63.4 & 77.8 & 63.2 & 78.9 & 63.8 & \textbf{1509} & 69.2 & 80.6 & \textbf{71.5} & 78.9 & 67.4 & \textbf{135}\\
\bottomrule
\end{tabular}}
\vspace{-3mm}
\caption{Comparison of the state-of-the-art methods under the “private detector” protocol on MOT17-test set and MOT17-val set.}
\vspace{-4mm}
\label{fig:mot17testval}
\end{table*}

\vspace{-4mm}
\paragraph{Analysis on different categories of SportsMOT.}
While the three SportsMOT categories share some common characteristics such as fast motion and similar appearance, they also have distinct features due to the different types of games. In this section, we analyze the results of our experiments on the three categories.

We first use the best HOTA metric from \cref{tab:alpha} to represent the \textit{overall difficulty} of a category. Based on this metric, we find that basketball videos (66.17) are the most difficult, followed by football (73.19), and finally volleyball (76.91).

Next, we consider the \textit{appearance-based} association (\ie $\alpha=0$ in \cref{tab:alpha}). We observe a notable gap between the HOTA metrics of the three categories, where basketball (26.94) has a much lower HOTA than volleyball (40.94) and football (65.65). Similarly, basketball remains the most difficult in the \textit{motion-based} association (\ie $\alpha=1$ in \cref{tab:alpha}), while the volleyball becomes the easiest.

We believe that the differences in difficulty arise from several factors, including the size of the game court and the degree of physical confrontation among players. For instance, basketball scenes are played on smaller courts and involve more physical contact between players than football scenes. This can lead to more occlusion and blur in basketball videos, making the association task more challenging than in football scenes.

\begin{table}[pt]
\resizebox{\linewidth}{!}{
\begin{tabular}{@{}ccc|ccccc@{}}
\toprule
IoU        & Motion     & Mix.    & HOTA$\uparrow$ & IDF1$\uparrow$ & AssA$\uparrow$ & MOTA$\uparrow$ & IDs$\downarrow$  \\ \midrule
\checkmark &            &            & 71.5 & 71.2 & 58.1 & 95.9 & 4329 \\
           &            & \checkmark & 64.2 & 63.9 & 48.7 & 91.1 & 25947 \\
\checkmark & \checkmark &            & 64.1 & 71.4 & 52.3 & 95.9 & 3089 \\
\checkmark &            & \checkmark & \textbf{73.8} & \textbf{74.4} & \textbf{61.6} & \textbf{96.6} & 3203 \\
\checkmark & \checkmark & \checkmark & 65.7 & 74.1 & 54.8 & 96.1 & \textbf{2469} \\ \bottomrule
\end{tabular}%
}
\vspace{-3mm}
\caption{Results of the ablation experiment on SportsMOT test set. IoU means computing IoU between detections and the last location of existing tracks for association, while Motion means using Kalman filter to predict the location of tracks. The models are trained on SportsMOT training and validation set.}
\vspace{-6mm}
\label{fig:ablation}
\end{table}

\vspace{-4mm}
\paragraph{Ablation study on MixSort.}
We ablate important components of our tracker (MixSort based on ByteTrack) including \textit{IoU}, \textit{Motion} (Kalman Filter) and \textit{MixSort} on SportsMOT test set. The results are presented in \cref{fig:ablation}.
Surprisingly, we found that simple IoU without using motion prediction outperformed IoU with motion prediction by a large margin (from 64.1 HOTA to 71.5 HOTA), indicating that the Kalman filter, which assumes linear motion models, performed poorly on SportsMOT, where the motion patterns are far more complex than in previous datasets.

Furthermore, we observed that MixSort played a crucial role in boosting the performance of the tracker significantly. By fusing the IoU and MixSort cues, our method achieved the best performance of 73.8 HOTA compared to 71.5 HOTA of simple IoU in our experiments.

\vspace{-4mm}
\paragraph{Comparison with SoccerNet.}
We conducted experiments to compare SoccerNet that focuses only on soccer scenes with SportsMOT. Results shown in \cref{fig:sportssoccer} suggest that SportsMOT is a challenging dataset with varying levels of difficulty across different sports categories. Specifically, basketball is proved to be the most difficult with the lowest HOTA of 60.8, while volleyball and football are relatively easier. 
MixSort obtains higher HOTA on SportsMOT than on SoccerNet. This is mainly because all the elements on the court are to be tracked in SoccerNet, leading to more false detections and much lower DetA (71.5 vs 78.8). However, it still obtains higher AssA on SoccerNet than on SportsMOT, in spite of the more false detections, which demonstrates that SportsMOT yields challenging association and is valuable for tracking in sports.
\begin{table}[pt]
\resizebox{\linewidth}{!}{
\begin{tabular}{@{}cc|ccccc@{}}
\toprule
                                                &            & HOTA$\uparrow$ & IDF1$\uparrow$ & AssA$\uparrow$ & MOTA$\uparrow$ & DetA$\uparrow$ \\ \midrule
\multicolumn{1}{c|}{\multirow{4}{*}{SportsMOT}} & overall    & 65.7 & 74.1 & 54.8 & 96.1 & 78.8 \\
\multicolumn{1}{c|}{}                           & basketball & \textbf{60.8} & \textbf{67.8} & \textbf{46.8} & 97.3 & 79.1 \\
\multicolumn{1}{c|}{}                           & volleyball & 72.5 & 87.0 & 66.8 & 96.5 & 78.7 \\
\multicolumn{1}{c|}{}                           & football   & 66.4 & 73.6 & 56.3 & 94.9 & 78.5 \\ \midrule
\multicolumn{2}{c|}{SoccerNet}                               & 62.9 & 73.9 & 55.5 & \textbf{87.8} & \textbf{71.5} \\ \bottomrule
\end{tabular}
}
\vspace{-3mm}
\caption{Results of MixSort-Byte on SportsMOT and SoccerNet test set. The models are trained on SportsMOT training and validation set and SoccerNet training set respectively and the hyper-parameters are the same. }
\label{fig:sportssoccer}
\end{table}

\vspace{-3mm}
\paragraph{Comparison with DanceTrack.}
We evaluate MixSort-Byte on the DanceTrack validation set and compare the results with that of ByteTrack as shown in \cref{fig:dance}. Unlike the results on SportsMOT where MixSort brings significant improvement, on DanceTrack the original ByteTrack performs better instead, with HOTA, MOTA and DetA metrics all higher than MixSort-Byte. This indicates that appearances in our proposed dataset SportsMOT are \textit{similar yet distinguishable}, while those in DanceTrack are much harder to distinguish. Therefore SportsMOT highlights both the motion-based and appearance-based associations.
\begin{table}[pt]
\vspace{-1mm}
\resizebox{\linewidth}{!}{%
\begin{tabular}{@{}c|ccccc@{}}
\toprule
             & HOTA$\uparrow$ & IDF1$\uparrow$ & AssA$\uparrow$ & MOTA$\uparrow$ & DetA$\uparrow$ \\ \midrule
ByteTrack    & \textbf{47.1} & 51.9 & 31.5 & \textbf{88.2} & \textbf{70.5} \\
MixSort-Byte & 46.7 & \textbf{53.0} & \textbf{31.9} & 85.8 & 68.6 \\ \bottomrule
\end{tabular}%
}
\vspace{-2.8mm}
\caption{Comparison of ByteTrack and MixSort-Byte on DanceTrack validation set. For MixSort-Byte, the fuse parameter $\alpha$ is 0.9, which results in the highest HOTA among \{0.6, 0.7, 0.8, 0.9, 0.95\}. The models are trained on DanceTrack training set.}
\vspace{-4mm}
\label{fig:dance}
\end{table}

\paragraph{Comparison Between MixSort and ReID models.}
To verify the effectiveness of the proposed MixSort with introducing a MixFormer-like model to model appearance association cues, we take experiments as in Table~\ref{tab:sup1}. 
We use the same ReID model as in DeepSORT and finetune it on our SportsMOT. 
We can see that, the HOTA, IDF1 and AssA of ByteTrack with ReID model are higher than that of original ByteTrack without ReID model, which demonstrate the importance of appearance-based association on the proposed SportsMOT.
Moreover, the proposed MixSort-Byte imporves ByteTrack with ReID model by 0.9, 1.9 and 1.4 on HOTA, IDF1 and AssA respectively.
This proves the superiority of MixSort's appearance model over the original ReID model, since it can extract more extensive and discriminative representations, and also allows more effective offline learning.

\begin{table}[pt]
\resizebox{\linewidth}{!}{%
\begin{tabular}{@{}c|ccccc@{}}
\toprule
             & HOTA$\uparrow$ & IDF1$\uparrow$ & AssA$\uparrow$ & MOTA$\uparrow$ & DetA$\uparrow$ \\ \midrule
ByteTrack    & 64.1 & 71.4 & 52.3 & 95.9 & 78.5 \\
ByteTrack+ReID    & 64.8 & 72.2 & 53.4 & \textbf{96.1} & \textbf{78.8} \\
MixSort-Byte &  \textbf{65.7} & \textbf{74.1} & \textbf{54.8} & \textbf{96.1} & \textbf{78.8} \\ \bottomrule
\end{tabular}%
}
\vspace{-3mm}
\caption{Comparison of ByteTrack, ByteTrack with ReID model and MixSort-Byte on SportsMOT test set. The ReID model is the same as in DeepSORT~\cite{bewley2016simple} and finetuned on SportsMOT. The models are trained on SportsMOT training set and the best results are shown in \textbf{bold}.}
\vspace{-4mm}
\label{tab:sup1}
\end{table}


\section{Conclusion}
In this paper, we have introduced~\emph{SportsMOT}, a large-scale multi-object tracking dataset in sports scenes.
SportsMOT is characterized with two key properties: 1) fast and variable-speed motion and 2) similar yet distinguishable appearance. 
We have empirically investigated several prevailing MOT trackers on the SportsMOT dataset.
We have also proposed a new MOT framework~\emph{MixSort}, introducing a MixFormer-like association module. 
Hopefully, SportsMOT can provide a platform for facilitating both sports analysis and multi-object tracking.


{\small
\bibliographystyle{ieee_fullname}
\bibliography{egbib}
}

\end{document}